\documentclass[preprint, review, 12pt]{elsarticle}

\usepackage{amssymb}
\usepackage{amsmath}
\usepackage{caption}
\usepackage{booktabs}
\usepackage[margin=1in]{geometry}
\usepackage{graphicx}

\usepackage{multirow}
\usepackage{rotating}
\usepackage{tabularx}
\usepackage{amsfonts}
\usepackage{mathtools}

\usepackage{subcaption}
\usepackage{float}
\usepackage{xcolor}
\usepackage{fancyhdr}		
\usepackage{enumitem}      
\definecolor{red}{rgb}{0.85,0,0}
\definecolor{blue}{rgb}{0,0,0.85}

\usepackage{lipsum}
\usepackage{makecell}
\usepackage{hyperref}
\hypersetup{ colorlinks,
    citecolor=blue,
    filecolor=blue,
    linkcolor=blue,
    urlcolor=blue
}
\journal{Int J Radiation Oncol Biol Phys}

\begin{document}

\begin{frontmatter}



\title{An Evidence-Driven Multi-Agent Compound AI System for Head and Neck Proton Radiotherapy Planning}

\author[1]{Qingqing Wang}
\ead{qiw058@health.ucsd.edu}
\author[1]{Liqiang Xiao}
\ead{leoxiaoliqiang@gmail.com}
\author[1,2]{Chang Chang}
 \ead{chang2@health.ucsd.edu}
\affiliation[1]{organization={Department of Radiation Medicine and Applied Sciences, University of California at San Diego},
             addressline={9500 Gilman Drive},
             city={La Jolla},
             postcode={92093},
             state={California},
             country={USA}}
 \affiliation[2]{organization={California Protons Cancer Therapy Center},
             addressline={9730 Summers Ridge Road},
             city={San Diego},
             postcode={92121},
             state={California},
             country={USA}}

\begin{abstract}
\noindent \textbf{Purpose:} Radiation treatment planning requires balancing competing objectives to satisfy clinical criteria while accommodating physicians' preferences.  
This study aims to develop an evidence-driven, multi-agent compound artificial intelligence (AI) system that integrates deep reinforcement learning (DRL)-based automated treatment planning with large language model (LLM)-based conversational plan refinement for proton pencil beam scanning (PBS) therapy. 

\noindent \textbf{Methods and Materials:} In the first stage, a policy network trained using Proximal Policy Optimization (PPO) generates multiple treatment plans without human intervention, from which one is selected as the target plan and the others serve as references for subsequent refinements. For challenging cases requiring further adjustments according to physicians' instructions, an LLM-based multi-agent system enables plan refinement through natural language interactions. The system comprises a planning agent that recommends evidence-supported adjustments, a chatbot providing a unified conversational interface, and an evaluation agent that retrieves clinical metrics and criteria and identifies constraint violations. To enhance domain-specific reasoning and decision-making, the planning agent, based on Qwen3-14B, is designed to justify its recommendations using evidence from previously generated plans and post-trained on task-specific planning data using supervised fine-tuning (SFT) and Group Relative Policy Optimization (GRPO). 
The framework was trained and evaluated using 21 patients with head and neck (H\&N) cancers, with five PPO-generated plans per patient. Of these, 50 plans were selected for instruction-guided refinement (42 for model training and eight for testing), with manually constructed adjustment instructions.

\noindent \textbf{Results:} Compared with the base LLM, the post-trained Qwen3-14B demonstrated improved instruction-following, information retrieval and domain-specific reasoning capabilities. The 
resulting adjustments more effectively improved user-prioritized objectives while limiting deterioration in competing objectives, yielding more clinically appropriate trade-offs in patient-specific plan refinement. The proposed compound AI system enabled conversational plan refinement through natural language instructions, providing a user-friendly planning platform and reshaping the planning workflow.

\noindent \textbf{Conclusions:} To the best of our knowledge, this is the first compound AI framework integrating DRL-based automated treatment planning with a domain-specific LLM agent post-trained for instruction-guided plan refinement, offering a promising approach to personalized proton PBS treatment planning. The proposed post-training strategy also provides a solution to reducing trial-and-error adjustment iterations, shortening planning time and lowering token costs for emerging agentic treatment planning systems.

\end{abstract}


\begin{highlights}
\item  A compound AI system integrating DRL-based automated planning with multi-agent conversational plan refinement is developed for proton PBS treatment planning.
\item A structured planning pipeline and an SFT-GRPO post-training strategy are proposed to develop an evidence-driven planning agent capable of recommending patient-specific adjustments supported by evidence from previously generated plans. 
\item Experimental results demonstrate that the proposed domain-specific post-training strategy enhances the planning agent's decision-making capabilities, enabling more effective plan adjustments aligned with user intentions while maintaining reasonable trade-offs among competing clinical objectives.
\end{highlights}


\end{frontmatter}



\section{Introduction}
\label{sec1}

Radiotherapy is a conerstone of cancer treatment worldwide, and the quality of a treatment plan can directly influence tumor control, treatment-related toxicity, and ultimately patients' survival and quality of life. Consequently, treatment planning has received considerable attention from the radiation oncology community. The planning process can be formulated as a multi-objective optimization problem, in which human planners iteratively adjust planning objectives and constraints for target volumes and surrounding organs at risk (OARs) to achieve an appropriate balance among competing clinical goals. This task becomes particularly challenging when OARs are in close proximity to, abut, or overlap with the tumor, making trade-offs between target coverage and normal-tissue sparing unavoidable.

In clinical practice, treatment plans often undergo multiple rounds of refinement before final approval, extending the planning process over several days. Such delays are undesirable because patient anatomy may change during this interval, potentially reducing the accuracy of dose delivery in a treatment modality that relies on precise spatial targeting. In addition, plan quality can vary substantially with factors such as planner experience, available planning time, and the capabilities of commercial treatment planning systems (TPSs). Suboptimal trade-offs among competing objectives may compromise either tumor coverage or normal-tissue sparing. Together, these challenges have created a strong need for efficient and reliable methods that can consistently generate high-quality treatment plans, thereby motivating the rapid development of automated treatment planning.

In last decades, automated treatment planning has been rapidly developed from knowledge-based planning (KBP)~\cite{B01, B02} and multi-criteria optimization (MCO)~\cite{B03} to virtual planners built with deep reinforcement learning (DRL)~\cite{B04, B05, B06, B07} and large language models (LLMs)~\cite{B08, B09, B10, B11, B12}. 
KBP estimates dose-volume histogram (DVH) curve bands from historical data for new patients by mathcing their anatomy to previous cases. MCO generates alternative plans by navigating the trade-off space along the approximate Pareto surface. However, the former collapses three-dimensional anatomy into two-dimensional DVH outputs, resulting in inverse optimization converging to inferior local minima and the necessity of mannual plan adjustments. The latter demands considerable manual effort on base plan generation, Pareto surface exploration and final plan selection~\cite{B12}. Both solutions fail to convert the time-consuming and expertise-demand treatment planning into desirable automatic process. 
Recently, DRL, a powerful approach to decision-making, has been increasingly explored for automated treatment planning. DRL-based methods typically enable policy networks adjustment intelligence through a trial-and-error learning process, using algorithms such as Q-learning~\cite{B04}, Soft Actor-Critic (SAC)~\cite{B05}, and Proximal Policy Optimization (PPO)~\cite{B06, B07}, etc. 
Although challenges remain in generalizing across diverse treatement sites and formulating reward functions that quantitatively capture the multifactorial assessment of plan quality, these methods have demonstrated promising progress and offer a viable approach to automated treatment planning.
Recent years have also  witnessed a surge in LLM-based applications in healthcare, particularly in high-stakes clinical decision-making tasks such as open-ended medical diagnostics~\cite{B13, B14, B15} and multidisciplinary treatment planning~\cite{B18}. Among these applications, radiotherapy treatment planning has emerged as an active area of research.
For example, Nusrat et al.~\cite{B09} proposed the Dose Optimization Language Agent (DOLA) for autonomous prostate radiatherapy planning, integrating chain-of-thought (CoT) prompting, retrieval-augmented generation (RAG) and reinforcement learning (RL) with Llama 3.1. Liu et al.~\cite{B10} developed GPT-RadPlan for volumetric modulated arc therapy (VMAT) planning, leveraging the multimodal model GPT-4V for plan evaluation and objective adjustment through in-context learning. Wei et al.~\cite{B08} employed prompt engineering to guide LLMs in adjusting optimization parameters for cervical cancer VMAT planning, incorporating feedback from medical physicists to improve planning decisions. More recently, Maniscalco et al.~\cite{B12} introduced a compound AI framework combining seven specialized LLM agents with a directive-conditioned 3D dose predictor. Through a structured multi-agent pipeline, these agents navigated the multi-objective optimization landscape by collaboratively performing clinical knowledge retrieval, DVH analysis, spatial dose assessment, plan compilation, quality assurance and tool execution. Collectively, these studies highlight the growing potential of LLMs to support complex planning decisions and advance automated radiotherapy treatment planning.

\begin{figure}[h]
\centering
\includegraphics[width=1.0\textwidth]{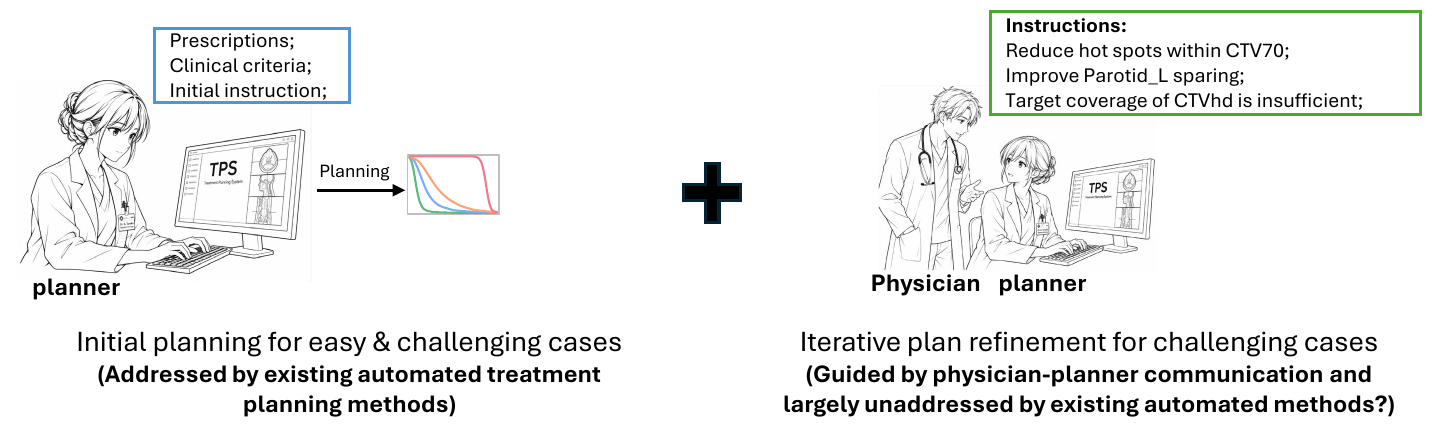}
\caption{Clinical practice of treatment planning and existing automatic treatment planning methods. }
\label{fig0}
\end{figure}

Although DRL and LLMs have significantly pushed the boundaries of automated treatment planning, DRL-based methods lack the flexibility to incorporate human instructions, while LLM-based approaches remain constrained by the limited domain-specific reasoning capabilities of base LLMs. Previous studies have demonstrated that existing LLMs may generate guideline-inconsistent or potentially harmful clinical recommendations~\cite{B22, B23, B24}. Although their outputs can provide complementary insights and support personalized decision-making, their treatment recommendations in precision oncology have not yet matched the quality and credibility of those made by human experts. Our experiments also indicate that, when given high-level instructions such as ``improve [OAR] sparing without compromising [TARGET VOLUME] coverage'', or ``reduce hot spots within [TARGET VOLUME] while maintaining its V100\% above 92\%'', base LLMs often recommend overly aggressive or ineffective adjustments, failing to appropriately balance improvements in the intended clinical objectives against potential deterioration in competing objectives. 
Moreover, medical LLM agents remain limited in their ability to dynamically adapt to changing patient conditions, further undermining their reliability in generating patient-specific treatment plans. 
Another limitation is that existing LLM-based agentic planning frameworks primarily focus on automated initial plan generation, overlooking the time-consuming iterative paln refinement guided by physician-planner communication, as illustrated in Figure~\ref{fig0}. This stage is particularly challenging when competing clinical objectives cannot be simultaneously satisfied, and physicians prioritize different objectives when making trade-offs. Furthermore, existing LLM-based automated plannig methods have primarily focused on photon-based radiotherapym such as VMAT and IMRT, while proton therapy planning remains relatively  underexplored.

In this work, we propose a multi-agent compound AI system for proton pencil beam scanning (PBS) radiotherapy planning. Specifically, a DRL-based policy network is first employed to automatically generate multiple treatment plans without human intervention. Subsequently, a conversational planning platform comprising three agents is developed to enable interactive plan refinement through natural language instructions. In particular, the planning agent is post-trained using SFT and GRPO to recommend patient-specific adjustments supported by evidence derived from previously generated plans, enhancing the reliability and interpretability of its recommendations. 
Compared with the base LLM, the post-trained model demonstrates improved instruction-following, information retrieval and clinical reasoning capabilities, enabeling more effective plan adjustments aligned with physicians' preferences while maintaining reasonable trade-offs among competing clinical objectives. To the best of our knowledge, this is the first compound AI framework specifically designed for proton PBS treatment planning that integrates automated initial plan generation with interactive, physician-guided plan refinement.

\section{Methods and Materials}
\label{sec2}

\subsection{Patient Cohorts}
This study was conducted under approval from the institutional review board of University of California San Diego. We retrospectively selected 21 bilateral or ipsilateral H\&N cases treated at the California Protons Cancer Therapy Center between 2017 and 2024. Across these cases, treatment courses consisted of 5--60 fractions delivered twice daily, with no more than five beams per plan. Each case contained one to four prescription levels spanning 4.0--74.2 Gy. Rather than redesigning the clinical setup, we retained the original prescriptions, beam angles, target definitions, and field-specific targets from the delivered plans. Moreover, following Wang et al.~\cite{B07}, proton spots were arranged on a hexagonal grid and dose was calculated using the Monte Carlo simulation. To ensure machine deliverability, the beam energy constraint of 70-240MeV and the minimal monitor units (MUs) of 3-300MUs per spot are enforced in inverse optimization.

\subsection{Automatic Treatment Planning Framework for Proton PBS}
\begin{figure}[h]
\centering
\includegraphics[width=1.0\textwidth]{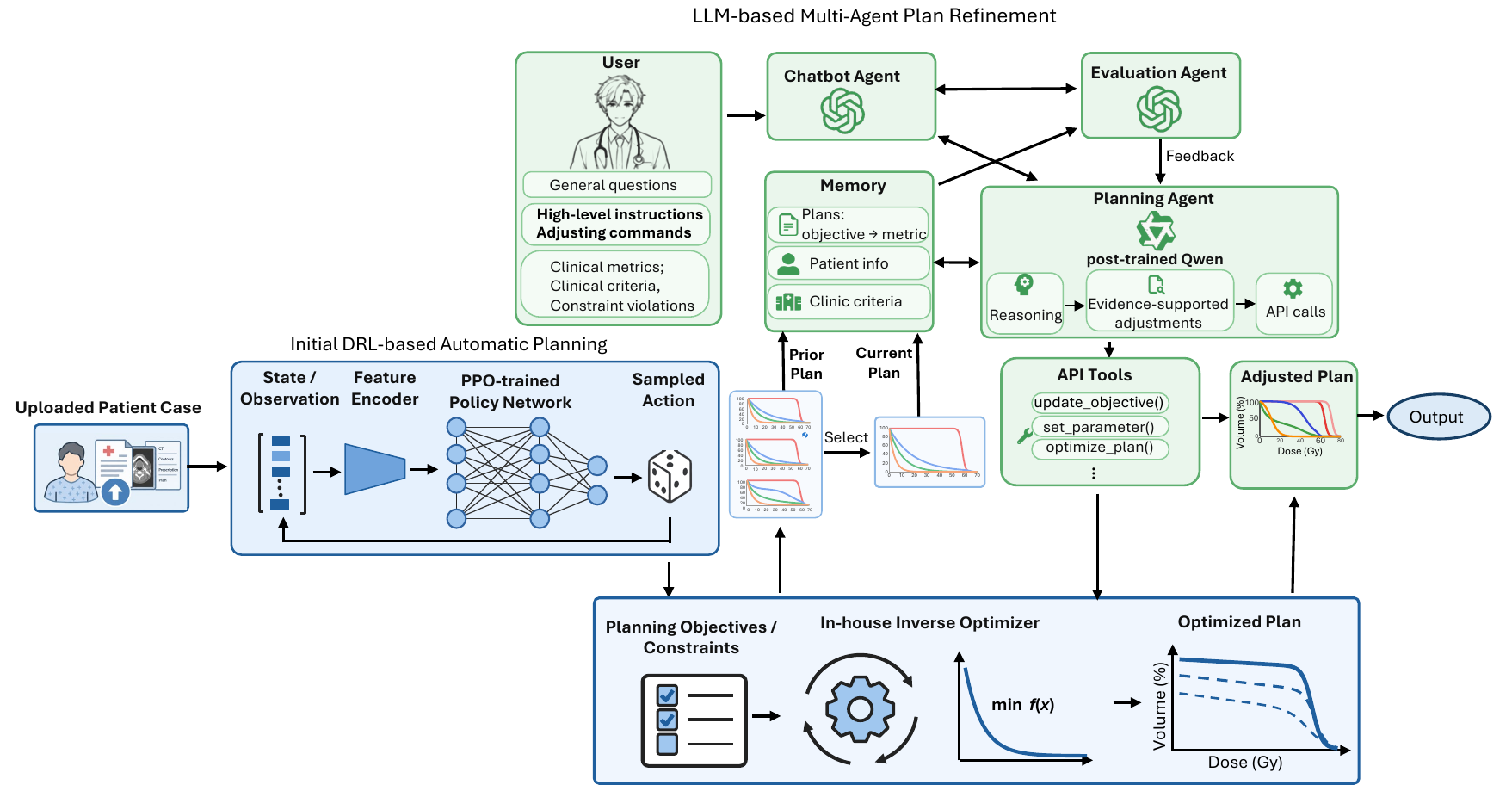}
\caption{Proposed evidence-driven multi-agent compound AI system for proton PBS treatment planning. }
\label{fig1}
\end{figure}

We perform treatment planning with the system depicted in Figure~\ref{fig1}, where an initial planning module built with an adjustment policy network is employed to automatically produce plans for uploaded patient cases without human intervention. The policy network is trained with the PPO framework proposed in Wang's previous work~\cite{B07}. Moreover, the advanced learn-to-optimize (L2O) inverse optimization engine designed in work~\cite{B07} is used as the backend of the planning system given its promissing effectiveness and efficiency. 
For easy cases, the initial planning module is likely to produce clinically accepted plans, while for challenging cases that physicians may give further refine instructions to, an LLM-based multi-agent planning module is designed to adjust a selected plan by reasoning over prior plans generated in the initial planning stage. Since the performance of the PPO-trained policy network and L2O inverse optimizer have been well demonstrated in Wang's work~\cite{B07}, we will focus on the multi-agent planning module, with a foucs on the post-training of the planning agent.

\subsection{LLM-based Multi-agent Planning System}
Physicians have different treatment preferences as they prioritize different objectives, thereby a conversational interface that enables human-in-the-loop planning is crucial. Recent developments in LLM-based agents provide a promising paradigm for facilitating nature language interaction between physicians and automated planning systems. Here, we propose an LLM-based multi-agent planning system as shown in Figure~\ref{fig1}. Specifically, the system consists of three collaborative agents: a chatbot agent, an evaluation agent and a planning agent. The first two agents are prompted GPT models, while the planning agent is a post-trained Qwen3-14B model. Functionally, the chatbot agent serves as the only user interface, responsible for answering general questions on radiation oncology and routing corresponding queries to other agents, e.g., routing high-level plan refine instructions and specific plan adjusting commands to the planner agent, and those relevant to clinical metrics, clinical criteria and constraint violations to the evaluation agent. The evaluation agent is designed to look up clinical requirements, evaluate plan qualities and check constraint violations by retrieving information from the memory module or calling pre-defined APIs. It also compares plans before and after adjustment and provides feedback to the planner agent. The planning agent is post-trained to adjust plans according to user's queries. For specific adjusting commands like ``reset objective parameters to [SPECIFIC VALUES]'', it invokes APIs directly and then confirms with the user. After being confirmed, the inverse optimization agent will be engaged to generate new plans. For high-level instructions like ``improve sparing of [OAR NAME]'', the planning agent recommends adjustments and a sequence of API calls that implement proposed adjustments by following a predefined planning workflow. It reasons over the objective-metric mapping of prior plans and justifies its recommendations with reliable evidences.

The system is empowered with a memory module and a set of API tools. The memory module stores patient information, clinical criteria and plan's objective-metric mapping information for Retrieval-Augmented Generation (RAG), while the API tools implement operations including DVH metric calculating, objective modifying, plan optimizing, etc for the evaluation and planning agents. As mentioned above, pre-trained LLMs may produce semantically meaningful adjustment recommendations, but the resulting plans may not be clinically acceptable because the simulated dose distributions frequently show unexpected degradations on competing objectives. 
Therefore, in this work, Qwen3-14B is post-trained on our planning-specific dataset to align its behavior with clinical plan-adjustment practices, thereby reducing the number of adjustment iterations and, consequently, LLM inference costs and overall planning time. 

Moreover, widely used LLM evaluation metrics are not directly applicable to our task because recommended adjustments that appear semantically correct may still be either too aggressive or too conservative to effectively produce a clinically acceptable plan. Instead, the performance of a planning agent should be evaluated based on the final simulation results of the adjusted plan, taking into account both the improvement of the targeted objectives specified in the user's instruction and the degradation of competing objectives that may not be explicitly mentioned. In addition, a single user instruction may contain conflicting goals that cannot be simultaneously satisfied. Therefore, the reasonableness of the trade-offs among targeted objectives should also be considered in the evaluation.
 
\subsection{Post-training of the Planning Agent}
When user proposes specific adjusting commands like ``reset the limit value of the [OBJECTIVE] for [OAR NAME] to [VALUE]'', the planning agent is allowed to invoke API calls directly without reasoning. However, when high-level instructions are given, the planning agent is prompted to follow a planning pipeline as below to recommend justified adjustments:
\begin{enumerate}
    \item Retrieve the objectives and corresponding clinical metrics from memory for all target volumes in the current plan;
    \item Retrieve the objectives and corresponding clinical metrics from memory for all targeted OARs and critical OARs whose clinical metrics fail to meet or are close to failing the clinical requirements;
    \item Compare the clinical metrics of the current plan and with those of prior plans, and select a prior plan that better reflects the user's intended trade-off while causing less degradation in competing objectives as the reference plan;
    \item Retrive the objectives and corresponding clinical metrics from memory for all target volumes, all targeted OARs and critical OARs in the selected reference plan;
    \item Recommend adjustments by comparing the objectives and clinical metrics of the current plan and the reference plan to satisfy user's intent, and suggest API calls that implement above adjustments;
    \item Output above API calls as the final response and ask the user to confirm proposed API calls.
\end{enumerate}
The first several steps are placed within a reasoning block enclosed by the ``<think>...</think>'' tag, and the last output step ensures that any modification to the current plan is confirmed by the user before execution. 

We collect training data using manually created queries and ground-truth responses, and post-train the Qwen3-14B thinking model using SFT and GRPO. SFT teaches the model what a good response looks like, while GRPO further optimizes which response is better in a group. Given GPU memory constraints, the widely used parameter-efficient fine-tuning method Low-Rank Adaptation (LoRA) is employed for both training stages. Moreover, the LLM-as-a-judge strategy is used to estimate rewards for the generated responses in GRPO.  

\subsubsection{Data Collection}
For each of the 21 cases, we generate five plans using the PPO-trained policy network from the initial planning stage. One plan is designated as the current plan, while the remaining plans are used as prior plans for reference. To improve the robustness of the LLM-based plannig agent by increasing the diversity of the training data, we select up to three different plans as the current plan for each case, and for each selected plan, we manually create one user query as plan-refinement instruction, resulting in a total of 50 training samples (42 for training and 8 for evaluation). 

For each sample, the ground-truth adjustment is first generated using a GPT model, which is prompted to recommend plan adjustments following the pipeline described above. For a small-scale, domain-specific training dataset, even a few low-quality samples can substantially affect model performance during post-training. Therefore, we carefully verify the correctness of the retrieved information and manually revise the generated response for each sample. In particular, we select appropriate reference plans and refine the recommended adjustments to ensure that they are consistent with the clinical planning practices of human planners.  

\subsubsection{Supervised Fine-tuning}
The supervised fine-tunning (SFT) stage aims to improve the model's instruction-following and format-following capabilities, enabling the planning agent to reason over prior plans according to the pipeline described above and recommend appropriate adjustments that refine the current plan toward user's intent. The think block incorporates retrieved objective settings and corresponding clinical metrics of the current and the reference plans, providing explicit evidence to support the recommended adjustments. 
By imitating well-structured expert planning behavior during SFT, the planning agent learns to retrieve more accurate and complete information, produce more consistent outputs and reduce invalid API calls, establish a strong initialization for subsequent reinforcement learning. 

The base LLM is fine-tuned for 10 epochs with a learning rate of $5\times 10^{-5}$ and batch size of 4 using the VeRL framework~\cite{B26}. Low-Rank Adaptation (LoRA)~\cite{B27}, a parameter-efficient fine-tuning method that learns compact low-rank updates to selected weight matrices while keeping the original model parameters frozen, is employed to enable efficient task-specific adaptation while largely preserving the general knowledge encoded in the pre-trained model.
Specifically, given a pre-trained weight matrix $\mathbf{W}_0 \in \mathbb{R}^{d \times k}$, LoRA introduces a trainable low-rank decomposition: \begin{equation} \Delta \mathbf{W} = \mathbf{B}\mathbf{A}, \end{equation} and adapts the weight matrix using \begin{equation} \mathbf{W} = \mathbf{W}_0 + \Delta \mathbf{W} = \mathbf{W}_0 + \mathbf{B}\mathbf{A}. \end{equation}
where $\mathbf{A} \in \mathbb{R}^{r \times k}$ and $\mathbf{B} \in \mathbb{R}^{d \times r}$ are trainable matrices, and $r \ll \min(d,k)$ is the rank of the adaptation. 
For an input representation $\mathbf{x}$, the forward pass becomes: \begin{equation} \mathbf{h} = \mathbf{W}_0 \mathbf{x} + \frac{\alpha}{r}\mathbf{B}\mathbf{A}\mathbf{x}, \end{equation} where $\alpha$ is a scaling factor controlling the strength of the LoRA update. For transformer-based language models, LoRA is commonly applied to attention projection matrices, such as query, key, value, and output projections. In our experiments, the LoRA rank $r$ and scaling factor $\alpha$ are set to 32 and 64, respectively.

\subsubsection{Group Relative Policy Optimization (GRPO)}
LLMs acquire substantial knowledge of oncology and treatment planning from large-scale pre-training corpora, and their strong reasoning capabilities enable them to recommend adjustments in generally appropriate directions. However, treatment planning is inherently case-sensitive. For example, both increasing the priority of an OAR objective and decreasing its dose limit value may improve OAR sparing, while different parameter values can lead to substantially different trade-offs with other competing objectives for different cases. Therefore, a key capbility of a planning agent is to determine \textbf{which adjustment is better for the current case.} From our experiments, despite access to prior plans as references, base LLMs may still recommend overly aggressive or ineffective adjustments. 
To further improve the agent's ability to distinguish among candidate adjustments and optimize case-specific planning decisions, we continue post-training the planning agent using GRPO~\cite{B28}, a reinforcement learning algorithm that optimizes the LLM policy without requiring a separate value model.  

Specifically, given a prompt $x$, the current policy $\pi_{\theta}$ samples a group of $G$ responses $\{y_i\}_{i=1}^{G}$, where $y_i \sim \pi_{\theta_{\text{old}}}(\cdot \mid x)$. Each response is evaluated by a reward function $r_i = r(x, y_i)$. Instead of estimating an absolute value baseline, GRPO normalizes rewards within the sampled group and computes a relative advantage: \begin{equation} A_i = \frac{r_i - \mathrm{mean}(\{r_j\}_{j=1}^{G})} {\mathrm{std}(\{r_j\}_{j=1}^{G}) + \epsilon}, \end{equation} where $\epsilon$ is a small constant for numerical stability. Based on this advantage, GRPO uses a clipped surrogate objective to constrain policy updates: \begin{equation} \mathcal{L}_{\text{GRPO}}(\theta) = -\frac{1}{G}\sum_{i=1}^{G} \min \left( \rho_i(\theta) A_i, \mathrm{clip}(\rho_i(\theta), 1-\epsilon_c, 1+\epsilon_c) A_i \right), \end{equation} where \begin{equation} \rho_i(\theta) = \frac{\pi_{\theta}(y_i \mid x)} {\pi_{\theta_{\text{old}}}(y_i \mid x)} = \prod_{t=1}^{T_i} \frac{\pi_{\theta}(y_{i,t} \mid x, y_{i,<t})} {\pi_{\theta_{\text{old}}}(y_{i,t} \mid x, y_{i,<t})}. \end{equation} 
In practice, a KL regularization term is often added to prevent the updated policy from drifting too far from a reference model $\pi_{\text{ref}}$. Therefore, the final loss function of GRPO is: 
\begin{equation} \mathcal{L}(\theta) = \mathcal{L}_{\text{GRPO}}(\theta) + \beta \, D_{\mathrm{KL}} \left( \pi_{\theta}(\cdot \mid x) \| \pi_{\text{ref}}(\cdot \mid x) \right), \end{equation} 
where $\beta$ controls the strength of the KL penalty. By comparing responses within the same group, GRPO provides a stable relative learning signal while avoiding the computational cost of training an additional critic or value network.

We train the supervised fine-tuned Qwen3-14B model using GRPO for 3 epochs with a learning rate of $1\times 10^{-5}$ and a batch size of 2. The KL loss coefficient $\beta$ is set to 0.001, and the response group size $G$ for each prompt $x$ is set to 3. The reward $r_i$ for each generated response $y_i$ is assigned by an LLM-as-a-Judge built on GPT-5.6, which evaluates the consistency between the generated response and the corresponding ground-truth response by estimating the clinical effects of their proposed API calls rather than relying on superficial similarity in wording. Specifically, the judge is instructed to evaluate consistency across four aspects: the adjusted structures, retrieved prescription doses, adjustment directions, and adjustment magnitudes, and to assign each generated response a score between 0 and 1.
By leveraging a strong language model as a judge for this difficult-to-verify planning task, complex outputs can be efficiently evaluated at scale while maintaining strong alignment with human judgments. 

\subsection{API Tools and Memory}
The tool-use capabilities of LLMs have substantially extended their problem-solving capacity beyond text generation.  In this work, we define a set of APIs that enable the evaluation agent to assess clinical plan metrics and the planning agent to modify planning objectives. 

\noindent\textit{For the evaluation agent:}
\begin{itemize}
\small
\item  api\_evaluator\_compute\_DMax/DMin/DMean(structureName):  computing the maximum dose, minimal dose, and average dose of a specified structure;
\item  api\_evaluator\_compute\_Dx/Dcc(structureName, value): computing the Dx and Dcc metrics of a specified structure;
\item  api\_evaluator\_compute\_Vx(structureName, value): computing the Vx metrics for a specified structure.
\end{itemize}
\noindent\textit{For the planning agent:}
\begin{itemize}
\small
\item  api\_planner\_modify\_objectives(structureName, metric, limitValue, weight):  modifying the constraint or priority of an existing objective for a specified planning structure;
\item  api\_planner\_inner\_optimization():  running inverse optimization to calculate spot MU values and generate a new dose distribution;
\item  api\_planner\_modify\_Rx(targetVolumeName, newRx): modifying the prescription dose level of a specified target volume; 
\item  api\_planner\_dose\_constraint(structureName, lower\_dose, upper\_dose, metric, weight, limitValue): cropping a subregion receiving dose in between [lower\_dose, upper\_dose] from an existing structure and constraint its dose by setting an objective on that subregion. This is used to detect the hot or cold areas from the target volumes or OARs;
\item  api\_planner\_volume\_constraint(structureName, volume, volume\_percentage, metric, weight, limitValue):  finding the most irradiated subvolume of a structure and constraint its dose by setting an objective on that subvolume.
\end{itemize}
    
To enable the proposed multi-agent planning system to generate patient-specific responses, a customized JSON-formatted memory is constructed to store treatment-related information, including beam configurations, patient information, clinical criteria, planning structures, prescriptions, and fractionation schemes. In particular, the memory stores objective-metric mappings from prior plans generated during the initial planning stage, which are crucial for the planning agent to infer appropriate adjustments and provide supporting evidence for its recommendations. 

\section{Experiments}
We qualitatively and quantitatively compare the planning performance of the base Qwen3-14B model and our post-trained Qwen3-14B model. First, we examine the models' natural-language outputs to assess their instruction-following, information retrieval, and reasoning capabilities. We then introduce a quantitative evaluation method to assess how well the plans adjusted by the base and post-trained Qwen3-14B models satisfy physicians' requirements, taking into account both improvements in the targeted clinical metrics and degradation in other competing objectives. As our primary goal is to evaluate the performance gains achieved through the proposed post-training strategy, both models are provided with the same prompts and memory to ensure a fair comparison. Finally, we present a conversational treatment planning system implemented using the proposed multi-agent framework to demonstrate how LLMs could reshape treatment planning workflows in the future.

\subsection{Comparison of LLM Outputs}
The LLM is instructed to recommend plan adjustments by following the predefined planning procedure. The intermediate steps within the think block provide supporting evidence for the recommendation, while the final API calls must be presented to the user for confirmation before execution. Retrieving metrics and objectives from the current and reference plans primarily tests the model's information retrieval capability, whereas adjustment recommendation rely more heavily on its reasoning ability. To illustrate how these capabilities change after post-training, we present a representative example with a challenging user query.

\begin{figure}[!htbp]
    \centering

    \begin{subfigure}{\linewidth}
        \includegraphics[width=\linewidth]{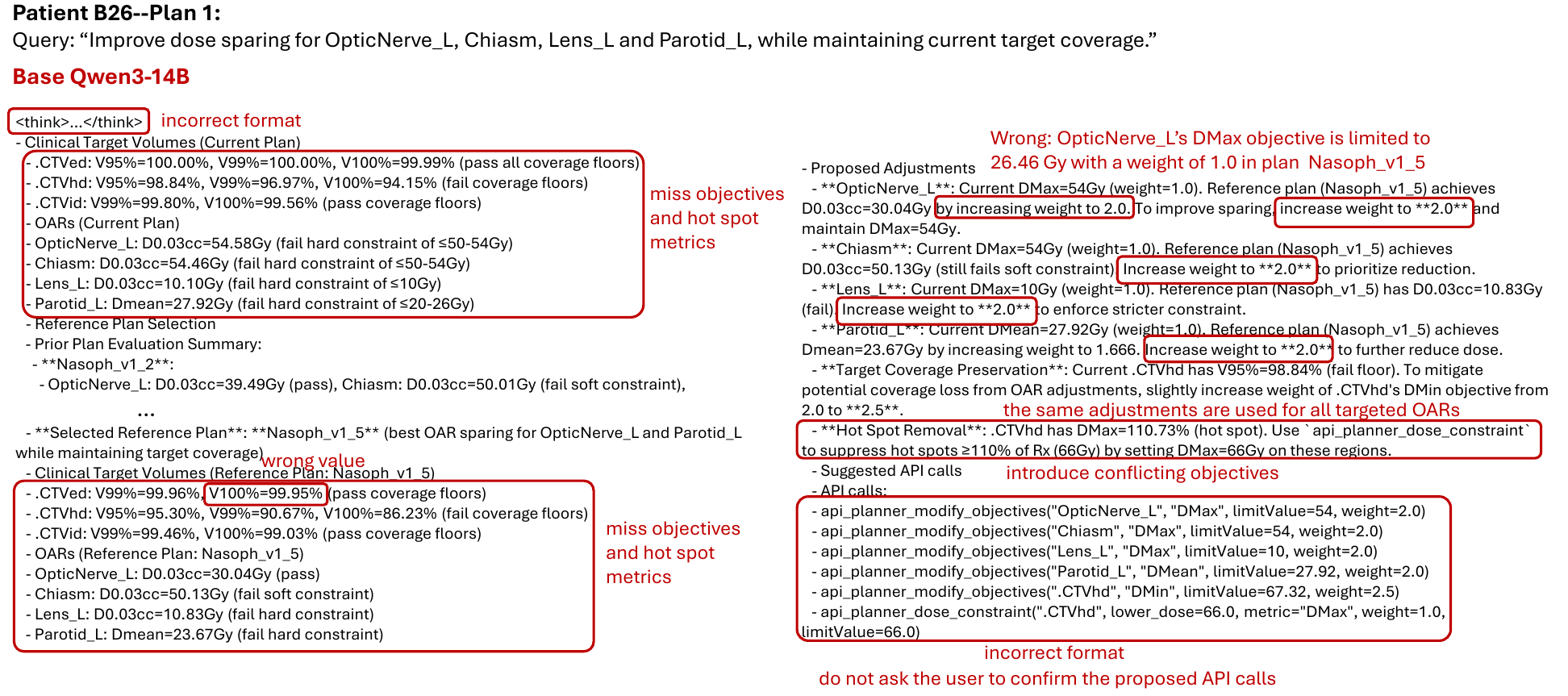}
    \end{subfigure}

    \vspace{1mm}

    \begin{subfigure}{\linewidth}
        \includegraphics[width=\linewidth]{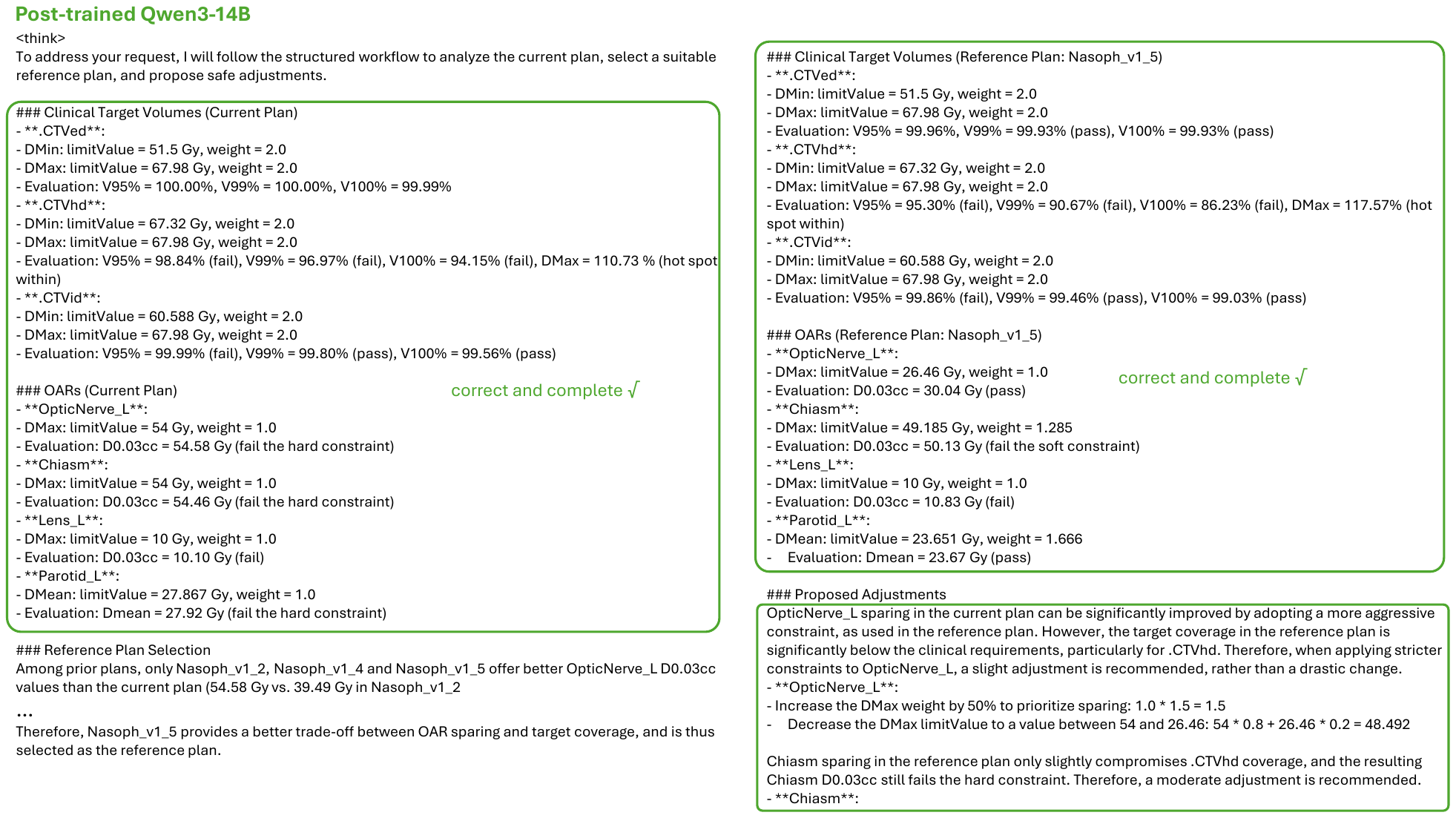}
    \end{subfigure}
    \begin{subfigure}{\linewidth}
        \includegraphics[width=\linewidth]{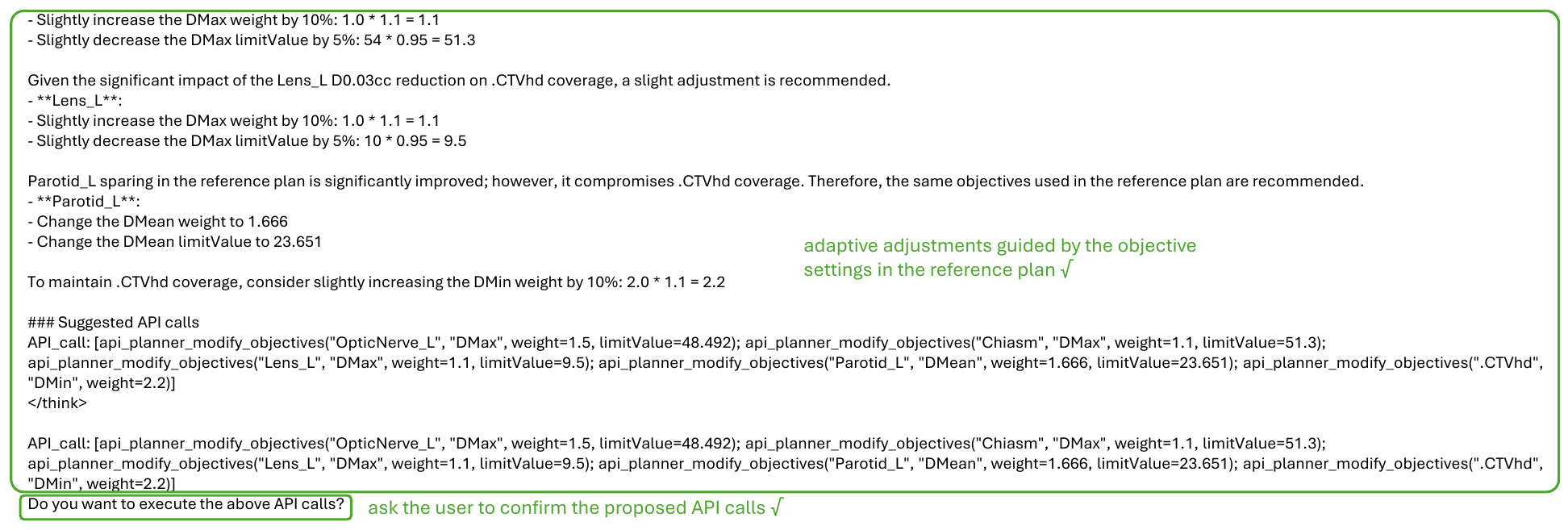}
    \end{subfigure}
    
    \caption{LLM output comparison of the base Qwen3-14B and post-trained Qwen3-14B.}
    \label{fig2}
\end{figure}

As shown in Figure~\ref{fig2}, although the instruction explicitly prescribes the reasoning procedure and requires API calls to be enclosed in brackets and separated by semicolons, the base Qwen3-14B model does not consistently follow the format, and instead tends to produce responses in a format resembling its pre-training behavior. It also fails to request user confirmation after proposing API calls and frequently omits key information during retrieval, particularly objective settings for planning structures. Information retrieval is especially challenging because each prior plan contains numerous clinical metrics and planning objectives, resulting in a large number of values stored in memory. The base model therefore often retrieves incorrect objective or metric values, which may further lead to hallucinated adjustment rationales. Nevertheless, the base model demonstrates basic treatment planning knowledge. It recognizes that OAR sparing can generally be improved by increasing objective weights and hot spots can be surpassed by adding an additional DMax objective to the hottest subregion. However, its recommendations are often generic rather than adapted to the settings of the current and the reference plans.  In Figure~\ref{fig2}, the base model increases the objective weights of all OARs to 2.0 without considering the objective settings of prior plans. The base model can also violate explicit planning constraints. For example, the DMin objective of .CTVhd is limited to 67.32 Gy, and therefore the DMax objective for any of its subregions should exceed 67.32 Gy. This rule is explicitly stated in the instruction. However, the base model introduces a DMax objective with a limit of 66 Gy for a subregion of .CTVhd receiving doses above 66 Gy, thereby creating conflicting objectives.

In contrast, the post-trained Qwen3-14B model follows the instructions substantially better. The think block and API calls are formatted as required, and the model explicitly asks the user to confirm the proposed API calls. It also correctly and completely retrieves the relevant objective and metric values from both the current and reference plans. 
In particular, the quality of the recommended adjustments improves substantially after post-training. The recommendations are guided by the objective settings of the reference plan, making them more clinically reasonable, while the supporting rationale is also more accurate and reliable. For example, for OpticNerve\_L, the base model states: ``Current DMax=54Gy (weight=1.0). Reference plan (Nasoph\_v1\_5) achieves D0.03cc=30.04Gy by increasing weight to 2.0. To improve sparing, increase weight to **2.0** and maintain DMax=54Gy''. However, the actual DMax objective in the reference plan is limited to 26.46 Gy with a weight of 1.0. By contrast, the post-trained model produces a more accurate and adaptive recommendation: ``OpticNerve\_L sparing in the current plan can be significantly improved by adopting a more aggressive constraint, as used in the reference plan. However, the target coverage in the reference plan is significantly below the clinical requirements, particularly for .CTVhd. Therefore, when applying stricter constraints to OpticNerve\_L, a slight adjustment is recommended, rather than a drastic change. **OpticNerve\_L**: Increase the DMax weight by 50\% to prioritize sparing: 1.0 * 1.5 = 1.5; Decrease the DMax limitValue to a value between 54 and 26.46: 54 * 0.8 + 26.46 * 0.2 = 48.492''.

Overall, the comparison shows that post-training substantially improves instruction following, particularly in response formatting and information retrieval. Reasoning performance is also improved, partly because the model has access to more complete and accurate retrieved information and, more importantly, because the adjustment recommendation policy is directly optimized through GRPO.

\subsection{Comparison of Recommended Adjustments}
When acting as a planning agent, an LLM should be evaluated based on how effectively its recommendations improve plan quality toward the user's intent, rather than on the semantic similarity between its natural language outputs and a ground truth response. Therefore, we apply the adjustments recommended by the LLM to the current plan, followed by inverse optimization and Monte Carlo dose calculation to generate the adjusted dose distribution. The LLM can then be evaluated using a quantitative framework that jointly considers: 1) how effectively the resulting plan improves the targeted objectives toward the physician’s intent, and 2) the extent to which these adjustments degrade other conflicting planning objectives. The comparison results of plans adjusted by the base and the post-trained Qwen3-14B are presented in Figure~\ref{fig3}. 
To reflect the diversity of plans to be adjusted by the proposed planning agent, we selected multiple plans with different trade-offs for challenging patients with bilateral cancer in the test set, such as Patient B25 and Patient B26.

For the first plan of Patient B25, the requirement is to improve Parotid\_R sparing without without increasing the dose to Parotid\_L or compromising target coverage. Compared with the baseline plan, (i.e., plan before adjustment), the base Qwen3-14B model reduces the mean dose to Parotid\_R from 28.8 Gy to 26.01 Gy, which still exceeds the clinical hard constraint of 26 Gy. In constrast, the post-trained Qwen3-14B model achieves a substantially greater reduction to 21.08 Gy, satisfying the hard constraint and approaching the soft constraint of 20 Gy. Both adjusted plans also slightly improve Parotid\_L sparing and target coverage, at the cost of slightly increased but clinically acceptable hotspots within CTV7000. Given that the primary objective is to improve Parotid\_R sparing, the post-trained model clearly outperforms the base model.

\begin{figure}[H]
    \centering
    \phantomcaption
    \label{fig:llm_adjust}

    \begin{subfigure}{\linewidth}
        \centering

        \includegraphics[width=\linewidth]{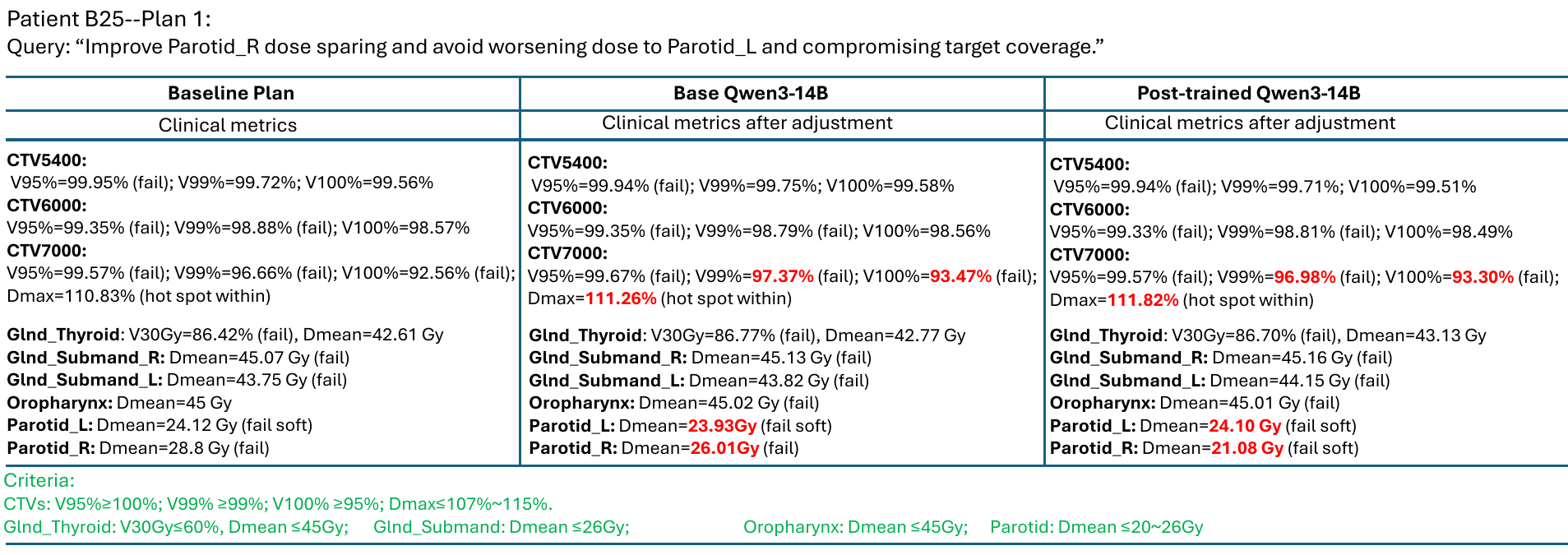}

        \vspace{1mm}

        \includegraphics[width=\linewidth]{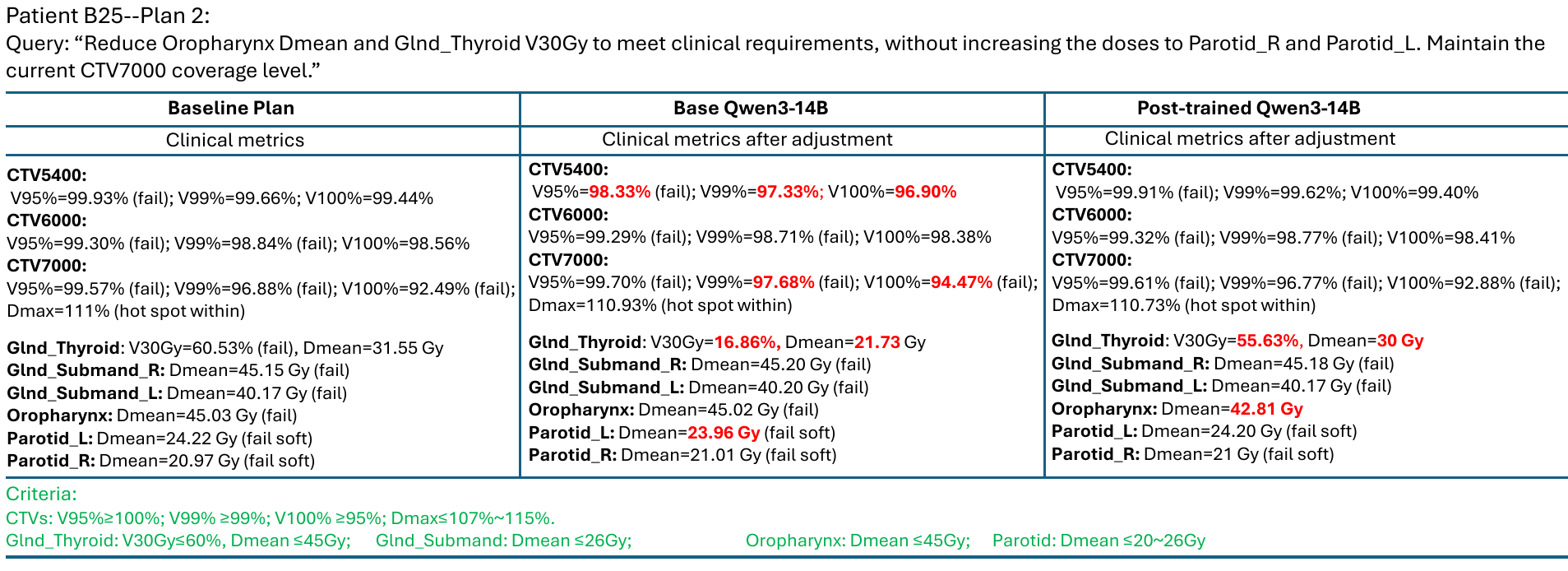}

        \caption{Patient B25 with bilateral cancer, with two plans selected for adjustment.}
        \label{fig:llm_b25}
    \end{subfigure}

\end{figure}
\begin{figure}[H]
    \ContinuedFloat
    \centering

    \begin{subfigure}{\linewidth}
        \centering

        \includegraphics[width=\linewidth]{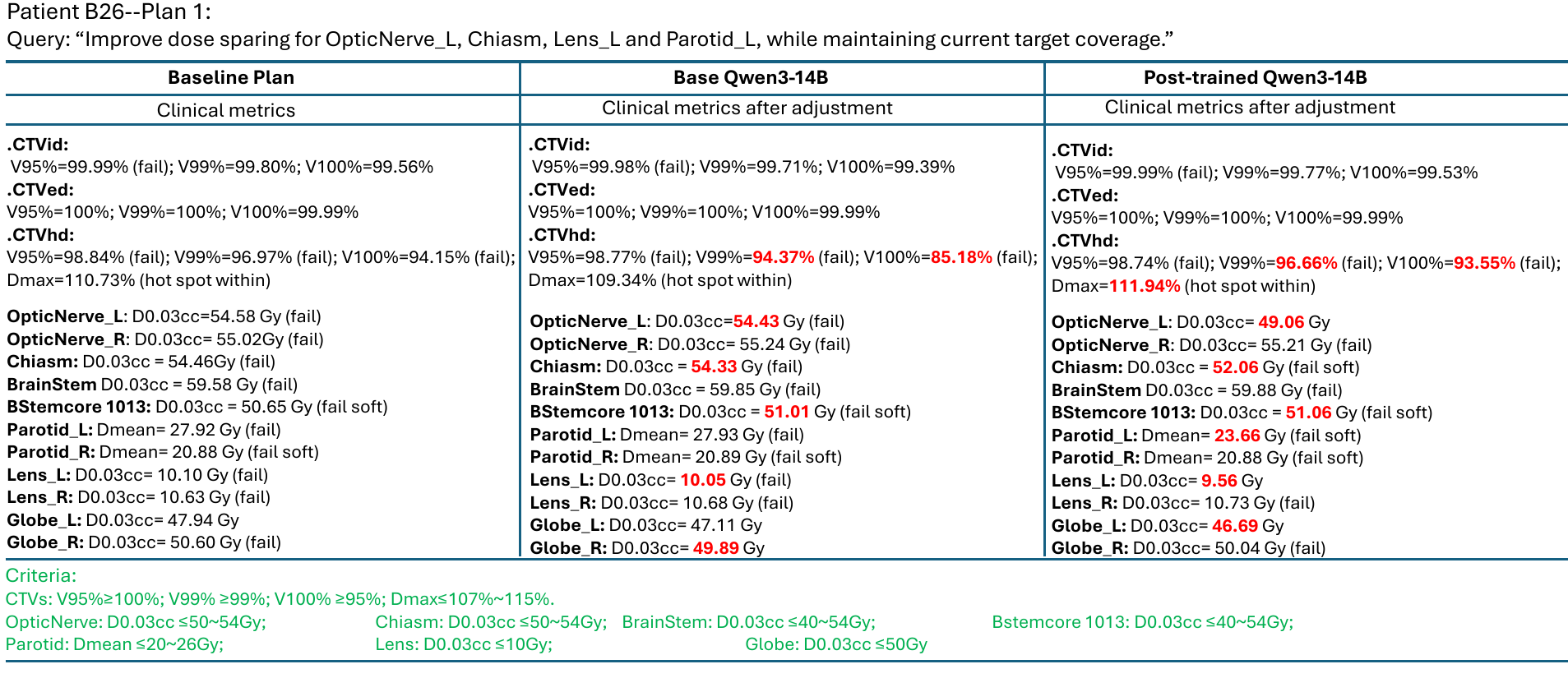}

        \vspace{1mm}

        \includegraphics[width=\linewidth]{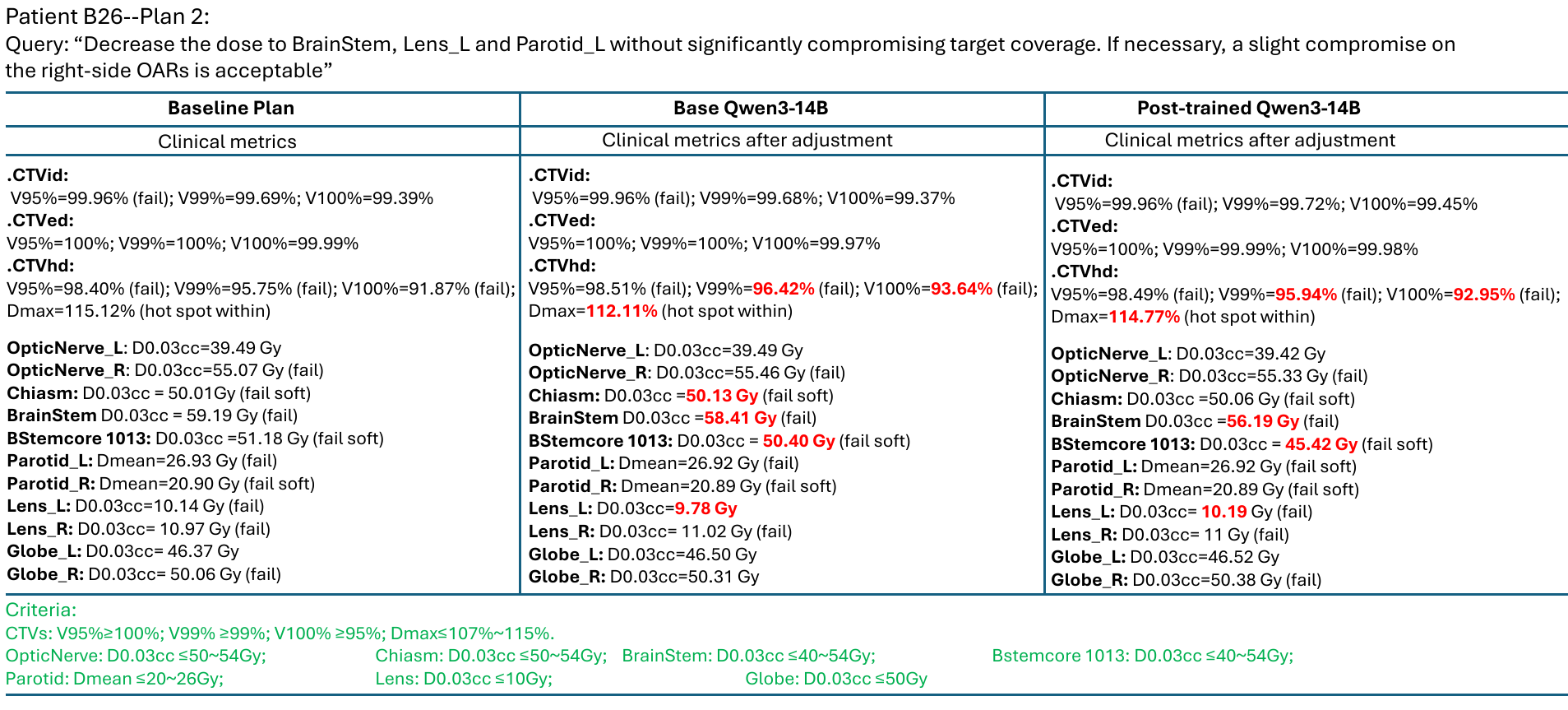}

        \vspace{1mm}

        \includegraphics[width=\linewidth]{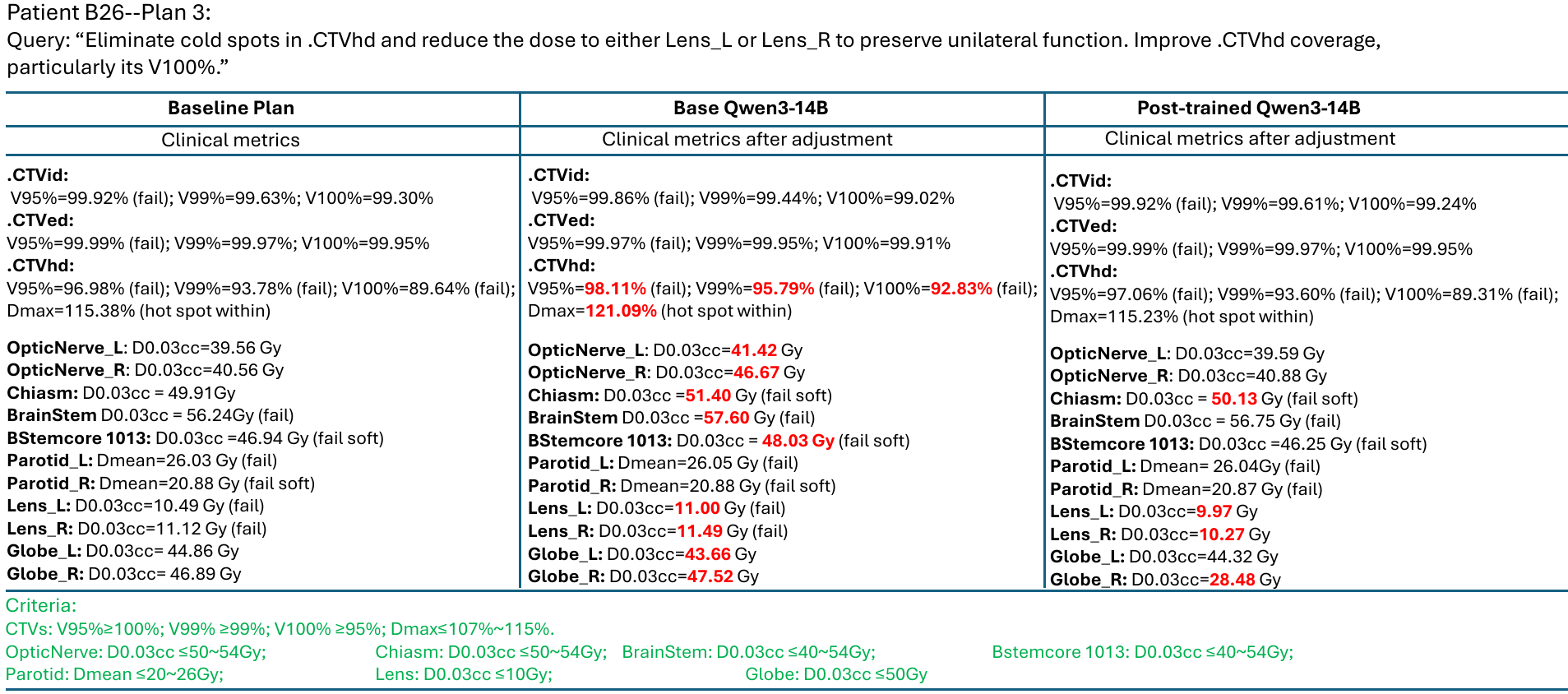}

        \caption{Patient B26 with bilateral cancer, with three plans selected for adjustment.}
        \label{fig:llm_b26}
    \end{subfigure}

\end{figure}
\begin{figure}[H]
    \ContinuedFloat
    \centering

    \begin{subfigure}{\linewidth}
        \centering

        \includegraphics[width=\linewidth]{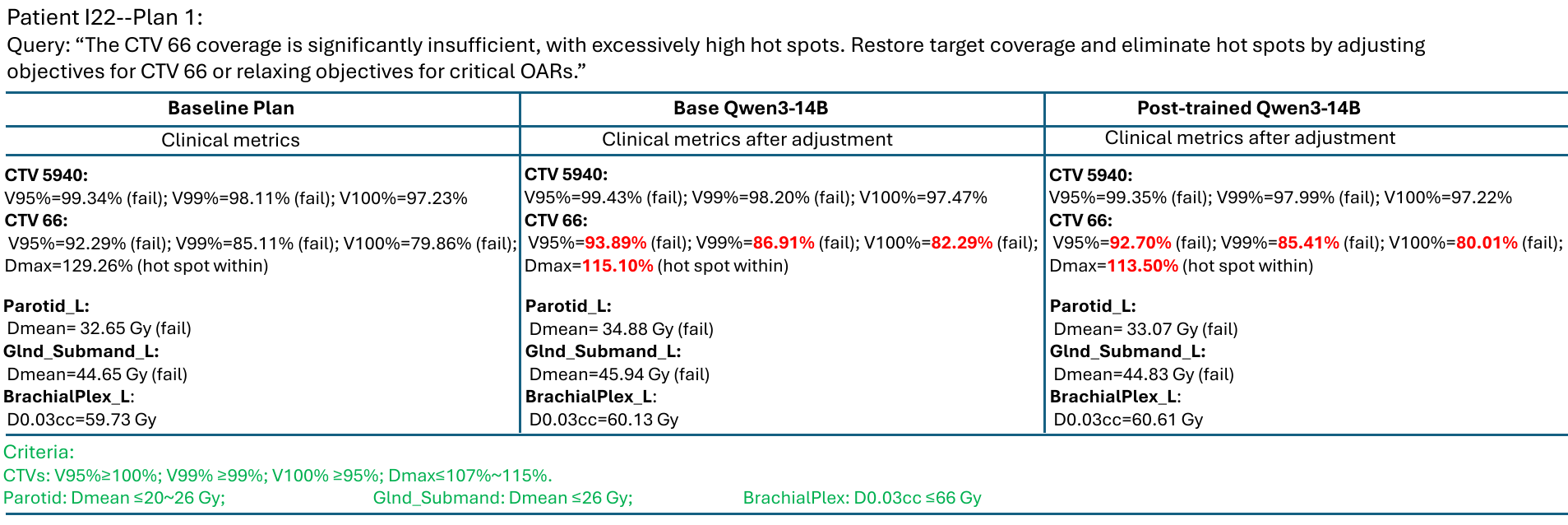}

        \vspace{1mm}

        \includegraphics[width=\linewidth]{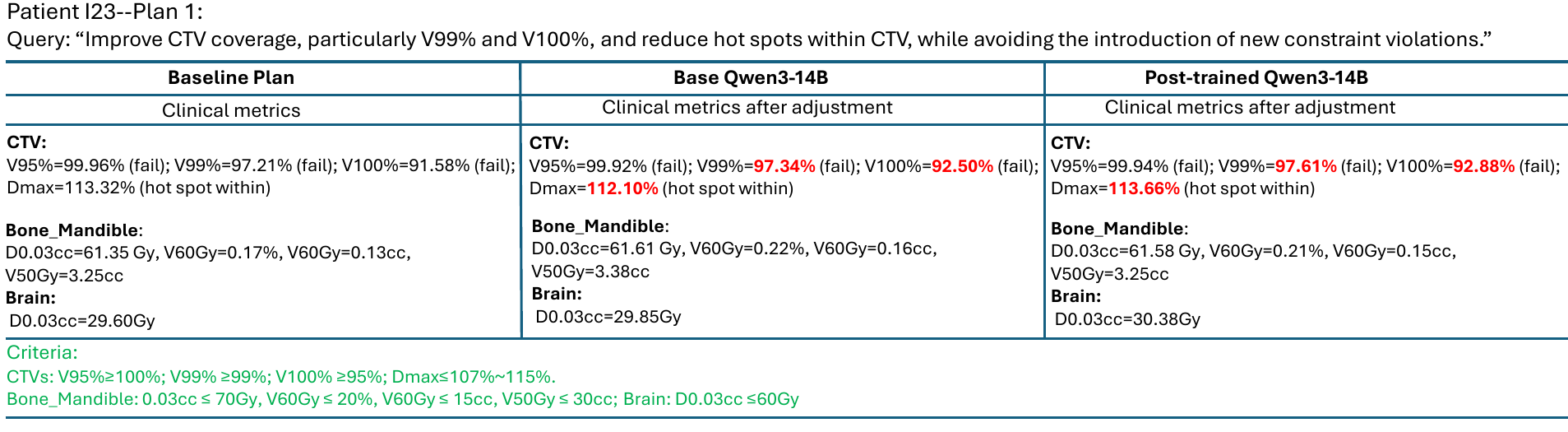}

        \vspace{1mm}

        \includegraphics[width=\linewidth]{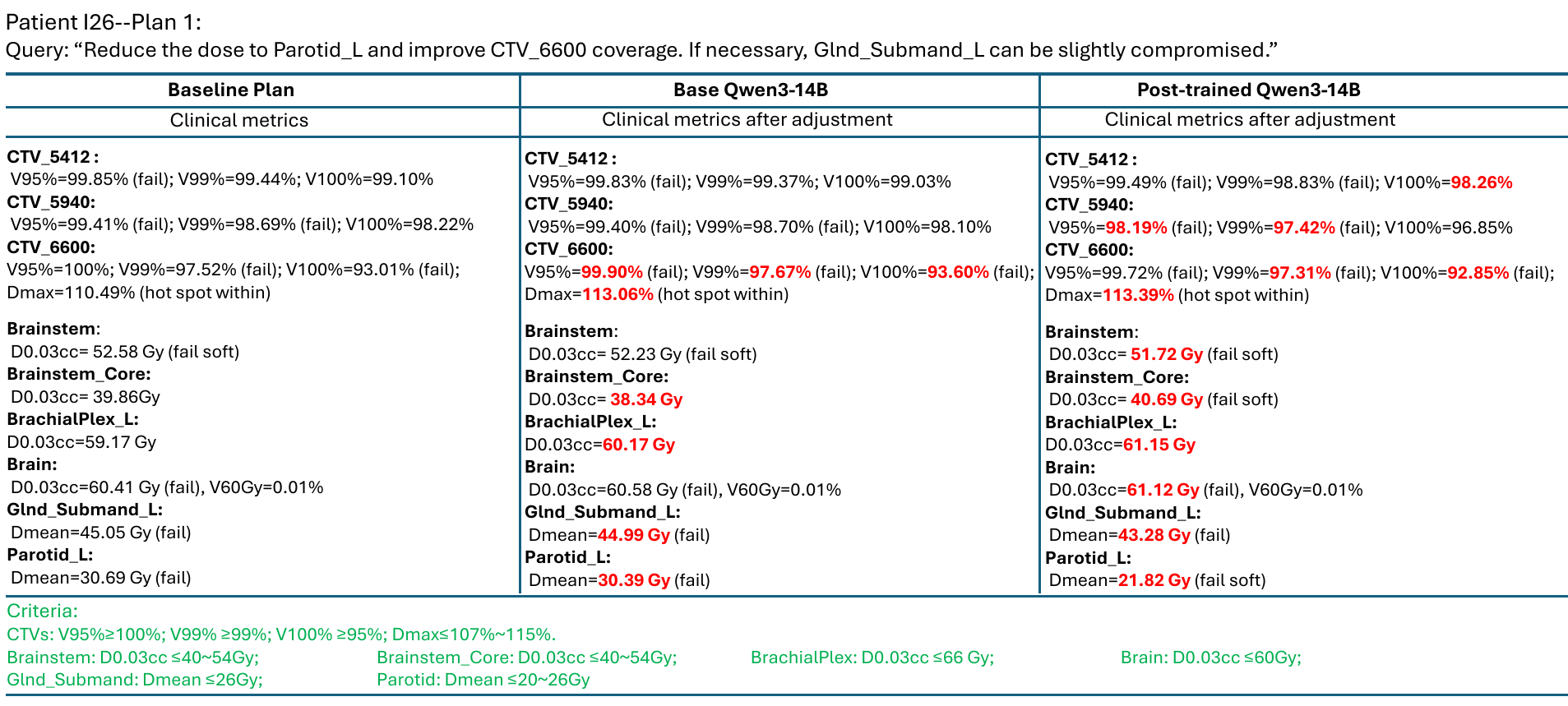}

        \caption{Patients I22, I23, and I26 with ipsilateral cancer.}
        \label{fig:llm_ipsilateral}
    \end{subfigure}

    \caption{Comparison of plans adjusted using recommendations from base Qwen3-14B and post-trained Qwen3-14B. OARs that meet the clinical criteria and show no noticeable changes before and after adjustment are not included in the comparison.}
    \label{fig3}
\end{figure}

The instruction for the second plan of Patient B25 is more challenging, requiring simultaneous improvements in Oropharynx and Glnd\_Thyroid sparing while maintaining target coverage and avoiding dose increases to either parotid gland.
The base model substantially reduces Glnd\_Thyroid $V_{30Gy}$ and mean dose to 16.8\% and 21.73 Gy, both well below the clinical constraints. However, Oropharynx sparing is almost unchanged before and after the adjustment (45.03 Gy vs. 45.02 Gy).  In constrast, the post-trained model reduces the mean doses to Oropharynx and Glnd\_Thyroid to 42.81 Gy and 30 Gy, respectively, both satisfying their 45 Gy constraints. It also reduces Glnd\_Thyroid $V_{30Gy}$ by approximately 5\% to a level meeting the clinical criterion. Neither adjusted plan produces noticeable changes in parotid doses, but their effects on target coverage differ substantially. The post-trained model maintains coverage across all three CTVs, whereas the base model significantly compromises CTV5400 coverage, reducing $V_{99\%}$ and $V_{100\%}$ by 2.33\% and 2.54\%, respectively, due to aggressive adjustments to Glnd\_Thyroid's objectives. Overall, the post-trained model achieves a more balanced trade-off among competing objectives, improving target objectives while better preserving target coverage.

For the first plan of Patient B26, the user requests improved sparing of OpticNerve\_L, Chiasm, Lens\_L, and Parotid\_L while maintaining target coverage. However, these objectives are inherently conflicting, making it difficult to satisfy all requirements simultaneously. The base model achieves only marginal improvements of less than 1 Gy for all four OARs, failing to meet the clinical criteria. More importantly, .CTVhd coverage deteriorates substantially, with $V_{100\%}$ decreasing from 94.15\% to 85.18\%, which is clinically unacceptable. In contrast, the post-trained Qwen3-14B model achieves substantially greater OAR sparing with minimal compromise in target coverage. Specifically, it reduces the $D_{0.03cc}$ of OpticNerve\_L, Chiasm, and Lens\_L by 5.52 Gy, 2.40 Gy, and 0.54 Gy, respectively, and the mean dose to Parotid\_L by 4.26 Gy, at the cost of only a 0.6\% reduction in .CTVhd $V_{100\%}$. The resulting OAR sparing satisfies the clinical hard constraints for all user-specified OARs.  Although hot spots within .CTVhd increase, the maximum dose of 111.94\% remains below the clinical hard constraint of 115\%. Overall, the post-trained model substantially outperforms the base model by achieving greater OAR sparing while better preserving target coverage. 
The second plan of Patient B26 requires improved sparing of another group of OARs, but neither model achieves the requested improvement in Parotid\_L sparing. Both models improve .CTVhd coverage and suppress hot spots within this target, but achieves only limited improvements in BrainStem and Lens\_L sparing. The base model achieves modest dose reductions in both OARs, decreasing their $D_{0.03cc}$ values by 0.78 Gy and 0.36 Gy, respectively. In contrast, the post-trained model prioritizes BrainStem over Lens\_L, reducing BrainStem $D_{0.03cc}$ by 3 Gy while achieving no improvement in Lens\_L sparing. This prioritization is clinically reasonable, as BrainStem is generally considered a more critical OAR than Lens\_L. Overall, the two models demonstrate comparable performance in this case: the post-trained model achieves a more clinically informed trade-off between competing objectives, whereas the base model successfully reduces Lens\_L $D_{0.03cc}$ below the clinical constraint. 
The third plan of Patient B26 suffers from insufficient target coverage, so the user requires the elimination of cold spots within .CTVhd and improved sparing of at least one lens to preserve unilateral visual function. The base model substantially improves target coverage, whereas the post-trained model maintains the baseline coverage level. However, the base model significantly compromises the sparing of other OARs, including OpticNerve\_L, OpticNerve\_R, Chiasm, and BrainStem. It also introduces more pronounced hot spots within .CTVhd, with a maximum dose of 121.09\% of the prescription dose, substantially exceeding the clinical hard constraint of 115\% of the prescription dose. Moreover, it increases the doses to both lenses, directly contradicting the user's request. In constrast, the post-trained Qwen3-14B model preserves the sparing of other OARs and does not introduce additional hot spots. It also reduces the doses to both lenses, bringing Len\_L $D_{0.03cc}$ below the clinical constraint. Overall, both models partially fulfill the user's requirements, but exhibit markedly different planning behaviors. The base model aggressively improves target coverage at the expense of other clinical objectives, whereas the post-trained model demonstrates greater restraint, achieving improved lens sparing while avoiding noticeable degradation in competing objectives.

To demonstrate how LLMs recommend personalized adjustments for patients with different clinical conditions, we apply the same query for Patient I22 and Patient I23, both with ipsilateral cancer, requesting improved target coverages and reduced hot spots. In the baseline plan of Patient I22, the mean doses to Parotid\_L and Glnd\_Submand\_L substantially exceed their clinical constraints of 26 Gy, while the maximum dose within CTV 66 reaches 129.26\% of the prescription dose, indicating severe hot spots. In contrast, Patient I23 exhibits better sparing across all OARs in the baseline plan, with hot spots already within the clinical constraint. 
For Patient I22, the base Qwen3-14B model improves CTV 66 $V_{95\%}$, $V_{99\%}$, and $V_{100\%}$ by 1.6\%, 1.8\%, and 2.43\%, respectively, while reducing its maximum dose to 115.10\% of the prescription dose. Despite these substantial improvements, the maximum dose still slightly exceeds the clinical hard constraint of 115\%. As a trade-off, the mean doses to Parotid\_L and Glnd\_Submand\_L are increased by 2.23 Gy and 1.29 Gy, respectively, worsening their already excessive doses. 
In constrast, the post-trained model achieves modest improvements in target coverage, while largely preserving the sparing of other OARs, avoiding further deterioration of critical structures that already violate clinical constraints. It also effectively suppresses hot spots, reducing the maximum dose within CTV 66 to 113.50\% of the prescription dose, thereby satisfying the clinical constraint.
For Patient I23, however, the post-trained model adopts a more aggressive adjustment strategy, achieving greater improvements in target coverage at the expense of increased doses to other OARs. Nevertheless, the resulting OAR doses remain within their clinical constraints, indicating that the model appropriately exploits the available margins for trade-offs.
These results demonstrate that the post-trained model adapts its adjustment strategy to patient-specific clinical conditions. It tends to make conservative trade-offs when competing OAR objectives already violate or approach their clinical limits, but adopts more aggressive adjustments when sufficient margins remain. In contrast, the base model lacks this adaptability. Overall, the post-trained model demonstrates better clinical awareness and more appropriate management of competing objectives, leading to more clinically reasonable and patient-specific treatment plan adjustments.

Patient I26 provides another example of how LLMs make different trade-offs when a single query contains requests that cannot be achieved simultaneously. Specifically, the user asks for improving Parotid\_L sparing and CTV\_6600 coverage. The base Qwen3-14B model achieves modest improvements in both Parotid\_L sparing and target coverage at the expense of increased doses to Brain and BrachialPlex\_L. In constrast, the post-trained Qwen3-14B model achieves a substantial improvement in Parotid\_L sparing, reducing its mean dose from 30.69 Gy to 21.82 Gy, at the cost of slightly compromised target coverage, with CTV\_6600 $V_{99\%}$ and $V_{100\%}$ decreasing by 0.21\%, and 0.16\%, respectively. This trade-off is clinically worthwhile, as target coverage remains close to the clinical requirement while Parotid\_L function can be preserved after adjustment.
Treatment planning inherently involves subjective decision-making since physicians may assign different priorities to competing objectives. Both models adjust plan toward user-specified goals without  excessively compromising other clinical objectives, potentially reflecting different physician preferences. Therefore, both adjusted plans can be considered clinically reasonable, despite their different trade-off strategies. 

Overall, the base LLM demonstrates basic treatment planning capabilities, but without domain-specific post-training, it may recommend overly aggressive or ineffective adjustments, and exhibits limited adapbility to patient-specific clinical conditions. When deployed as a planning agent, such a suboptimal LLM may require more iterations to achieve a physician-satisfactory and clinically acceptable plan, resulting in prolonged planning time and higher token costs. Our experimental results demonstrate that the predefined evidence-supported planning pipeline and the proposed post-training strategy effectively enhance the LLM's planning capabilities, better aligning its behavior with clinical practice and enabling more adaptive, physician-guided plan adjustment.

\subsection{Conversational Planning Using Proposed Multi-agent framework}
\begin{figure}[!htbp]
    \centering

    \begin{subfigure}{\linewidth}
        \includegraphics[width=\linewidth]{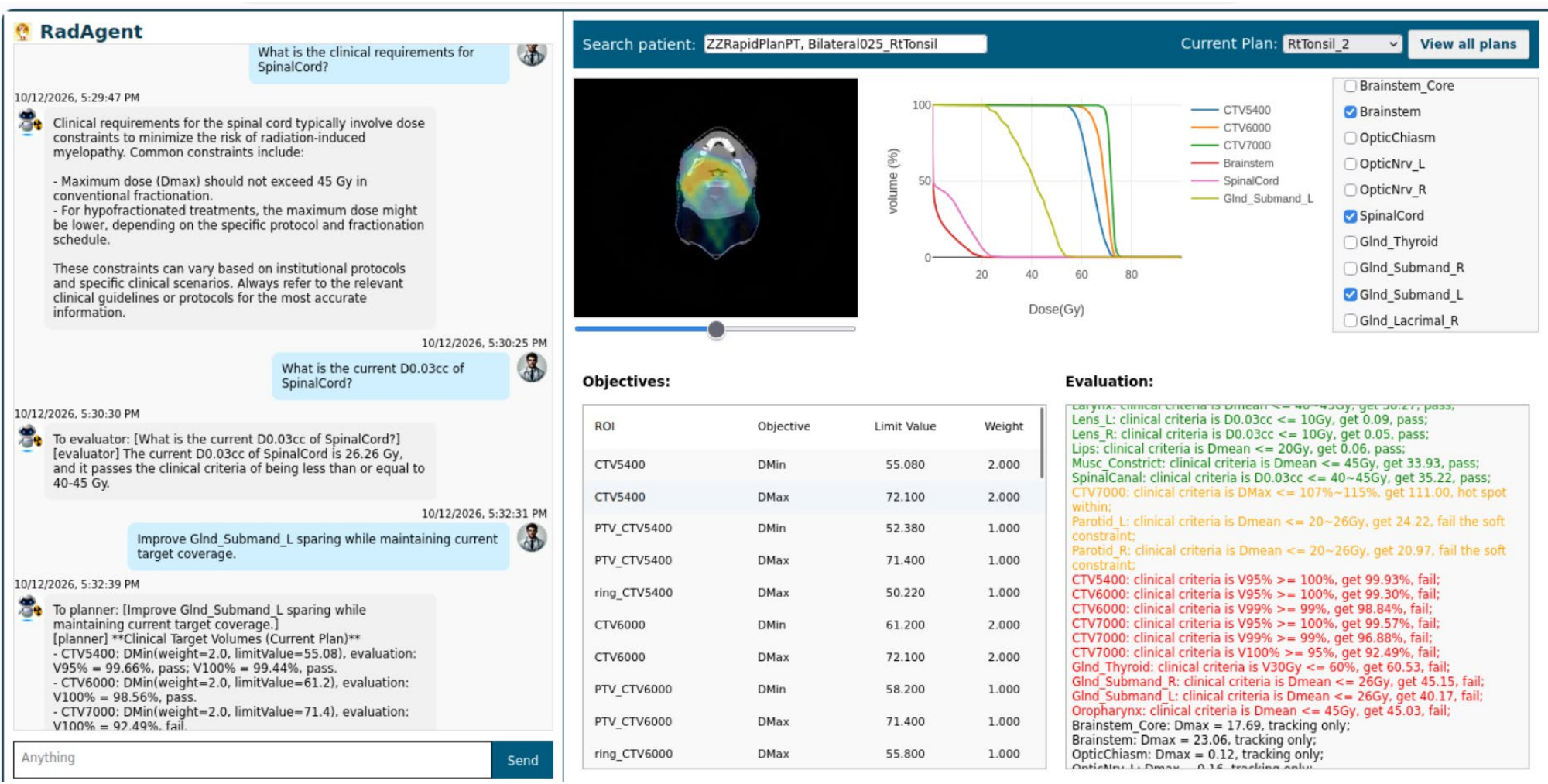}
    \end{subfigure}

    \vspace{2mm}

    \begin{subfigure}{\linewidth}
        \includegraphics[width=\linewidth]{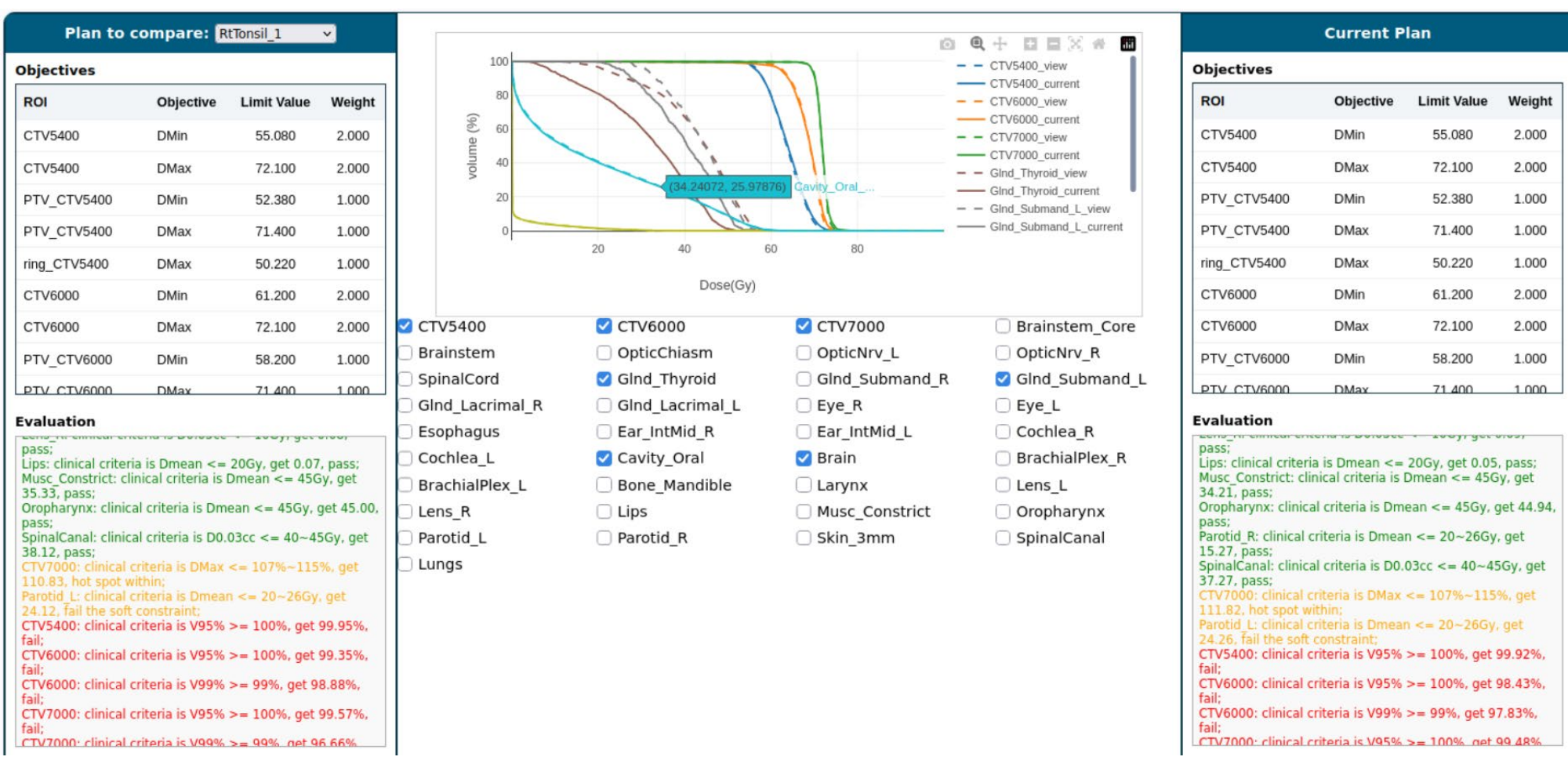}
    \end{subfigure}
    
    \caption{Conversational treatment planning using proposed multi-agent framework.}
    \label{fig4}
\end{figure}

LLMs are transforming workflows across various domains, as exemplified by the rapid rise of vibe coding in software development and AI-assisted diagnosis in healthcare. These advances have the potential to improve efficiency, reduce time and labor costs, and alleviate workforce shortages. In this work, we demonstrate how LLMs faciliate treatment planning by developing a novel, user-friendly conversational planning system built upon the proposed multi-agent framework, as illustrated in Figure~\ref{fig4}.

For relatively simple cases, the initial DRL-based automatic planning module can generate clinically acceptable plans with a high success rate without human intervention. Users only need to compare the generated plans and select the one that offers the best trade-offs among competing clinical objectives. For more challenging cases, users can provide additional instructions to further refine the plans.
Users, including physicians and dosimetrists, can instruct the planning agent to modify planning objective parameters through specific adjustment commands or request plan modifications using high-level instructions. Users can also inquire about clinical criteria, treatment plan metrics, clinical constraint violations, and general oncology-related topics through natural language interactions. 
Overall, by mimicking the workflow of an experienced treatment planner in both the initial and subsequent planning stages, and enabling users to efficiently refine treatment plans with the assistance of a post-trained planing agent, the proposed system has the potential to substantially reduce the workload of treatment planners and the overall planning time.

\section{Discussion}
Rapid advances in LLMs have the potential not only to transform treatment planning workflows but also, more importantly, to improve the quality of generated treatment plans by supporting better planning decisions. To fully realize this potential, developing domain-specific LLMs tailored to treatment planning is crucial.
In this work, we post-trained a planning agent for a multi-agent treatment planning framework and demonstrated the effectiveness of the proposed plannig pipeline and post-training stategy in enhancing the LLM's ability to recommend patient-specific plan adjustments.
By enabling users to guide plan refinement through high-level instructions, the proposed planning agent addresses a key limitation of existing automatic treatment planning systems: their limited ability to interact with physicians and refine treatment plans according to clinical preferences in a manner similar to human planners. Moreover, the planning agent can justify its recommendations using evidence derived from the planning objectives and clinical metrics of previously generated plans, improving the interpretability and reliability of the proposed adjustments.
By integrating the planning agent with a DRL-based automatic planning module capable of generating high-quality plans without human intervention, the proposed multi-agent framework supports both initial automatic planning and subsequent interactive plan refinement, addressing two essential stages of the clinical treatment plannig workflow.

This pilot study represents an initial effort to bridge automatic plan generation and instruction-guided plan refinement. While the proposed framework explores the key components of such an integrated planning system, further development is needed to enhance its capabilities, improve user experience, and facilitate its translation into clinical practice.

First, prompt-guided planning structure creation is essential for improving system flexibility. In this work, the APIs api\_planner\_dose\_constraint() and api\_planner\_volume\_constraint() enable the planning agent to extract subregions from existing planning structures based on specified dose ranges or highly irradiated volumes. However, users may also need to create auxiliary planning structures before plan optimization through natural language instructions, such as ``Subtract the overlapping region between the brainstem and CTV from the brainstem'', or ``Create a planning structure encompassing the portion of the brainstem closest to the tumor to facilitate localized dose sparing''. Wang et al~\cite{B06, B07} generated auxiliary planning structures using empirical rules, but more flexible, prompt-guided approaches could provide greater convenience. One possible solution is to develop a set of APIs supporting geometric and Boolean operations on anatomical structures, allowing the planning agent to sequentially execute these operations to generate the desired structures. Alternatively, the framework could be integrated with an advanced prompt-guided medical image segmentation model to enable more flexible structure generation.
Second, automatic beam angle selection, an important component of treatment planning, has not been addressed in this work.
Ferrante et al~\cite{B11} demonstrated the potential of pretrained vision-language models (VLMs), such as GPT-4V, for determining gantry angles in photon radiotherapy. These findings suggest that VLM-based approaches may also be promising for beam angle selection in proton PBS therapy. Incorporating automatic beam angle selection into the proposed framework could further improve planning automation and reduce the need for manual intervention.
Third, robust optimization, a critical component of clinically acceptable proton treatment planning, was not incorporated into the current framework. One major challenge is the computational cost associated with repeated Monte Carlo dose calculations. GPU-accelerated Monte Carlo dose calculation offers a promising approach to improving computational efficiency. Therefore, integrating GPU-accelerated Monte Carlo dose calculation and robust optimization into the proposed framework will be an important direction for future research.
Finally, expanding the training dataset and incorporating explicit supervision of intermediate decision-making steps, particularly reference plan selection, could further enhance the planning agent's ability to generate accurate and consistent recommendations. 

\section{Conclusion}
In this work, we developed a compound AI system  integrating DRL-based automated treatment planning with multi-agent conversational plan refinement for proton PBS treatment planning. To enhance the decision-making capabilities of the core planning agent, we proposed a structured planning pipeline and an SFT-GRPO post-training strategy to enable evidence-driven, patient-specific plan adjustments. Experimental results demonstrated that post-training substantially improved the agent's instruction-following, information retrieval, and clinical reasoning capabilities, enabling more effective adjustments aligned with user intentions while maintaining reasonable trade-offs among competing clinical objectives.
This study represents an initial effort to integrate automated initial planning with physician-guided interactive plan refinement for proton PBS treatment planning. Beyond enabling a more flexible and user-friendly planning workflow, our findings highlight the importance of domain-specific post-training in improving the decision-making capabilities of LLM-based planning agents. The proposed framework provides a promising foundation for developing more intelligent, interactive, and clinically adaptable agentic treatment planning systems.


\begin{thebibliography}{00}
\bibitem{B01}
Zhang J, Ge Y, Sheng Y, Wang C, et al.
\newblock {Knowledge-based tradeoff hyperplanes for head  and neck treatment planning.}
\newblock Int. J. Radiat. Oncol. * Biol. * Phys. {\bf 106}, 1095-103 (2020).

\bibitem{B02}
Momin S, Fu Y, Lei Y, Roper J, et al.
\newblock {Knowledge-based radiation treatment planning: a  data-driven method survey.}
\newblock J. Appl. Clin. Med. Phys. {\bf 22}, 16-44 (2021).

\bibitem{B03}
Miguel-Chumacero E, Currie G, Johnston A, Currie S.
\newblock {Effectiveness of multi-criteria optimization-based trade-off exploration in combination with RapidPlan for head \& neck radiotherapy planning.}
\newblock Prog. Med. Phys. {\bf 13}, 229 (2018).

\bibitem{B04}
Shen C, Gonzalez Y, Klages P, et al.
\newblock {A hierarchical deep reinforcement learning framework for intelligent automatic treatment planning of prostate cancer intensity modulated radiation therapy.}
\newblock Phys. Med. Biol. {\bf 66}, 134002 (2021).

\bibitem{B05}
Yang D, Wu X, Li X, et al.
\newblock {Automated treatment planning with deep reinforcement learning for head-and-neck (HN) cancer intensity modulated radiation therapy.}
\newblock Phys. Med. Biol. {\bf 70}, 015010 (2025).

\bibitem{B06}
Wang Q, Chang C.
\newblock {Automating the optimization of proton PBS treatment planning for head and neck cancers using policy gradient-based deep reinforcement learning.}
\newblock Med. Phys. {\bf 52}, 1997-2014 (2025).

\bibitem{B07}
Wang Q, Xiao L, Chang C.
\newblock {A learning-driven automatic planning framework for proton pencil beam scanning treatments of heand and neck cancers.}
\newblock Int. J. Radiat. Oncol. * Biol. * Phys. (2026).

\bibitem{B08}
Wei S, Hu A, Liang Y, Yang J, et al.
\newblock {Feasibility study of automatic radiotherapy treatment planning for cervical cancer using a large language model.}
\newblock Radiat. Oncol. {\bf 20}, 77 (2025).

\bibitem{B09}
Nusrat H, Luo B, Hall R, Kin J, et al.
\newblock {Autonomous radiotherapy treatment planning using DOLA: a privacy-preserving LLM-based optimization agent.}
\newblock arXiv:2503.17553 (2025).

\bibitem{B10}
Liu S, Pastor-Serrano O, Chen Y, Gepaulchan M, et al.
\newblock {Automated radiotherapy treatment planning guided by GPT-4Vision.}
\newblock Phys. Med. \& Biol. {\bf 70}, 155002 (2025).

\bibitem{B11}
Ferrante M, Carosi A, D'Angelillo R, Toschi N.
\newblock {Transforming multimodal models into action models for radiotherapy.}
\newblock Appl. Med. Artif. Intell. {\bf 15384}, 41-53 (2025).

\bibitem{B12}
Maniscalco A, Park Y, Domal S, Lin M, et al.
\newblock {Autonomous radiotherapy planning via agentic orchestration using a multimodal TPS-integrated compound AI platform.}
\newblock Mach. Learn.: Health {\bf 2}, 025005 (2026).

\bibitem{B13}
Zoller N, Berger J, Lin I, Fu N, et al.
\newblock {Human-AI collectives most accurately diagnose clinical vignettes.}
\newblock Proc. Natl. Acad. Sci. U.S.A. (PNAS) {\bf 122}, 2426153122 (2025).

\bibitem{B14}
Liu P, Bansal S, Dinh J, Pawar A, et al.
\newblock {MedChat: a multi-agent framework for multimodal diagnosis with large language models.}
\newblock  arXiv:2506.07400v2,  (2025).

\bibitem{B15}
Palepu A, Dhillon V, Niravath P, Weng W, et al.
\newblock {Exploring large language models for specialist-level oncology care.}
\newblock  arXiv:2411.03395v1,  (2024).



\bibitem{B18}
Kim J, Lee B, Sohn K, Kim D, Lee Y, et al.
\newblock {Evaluation of multi-agent LLMs in multidisciplinary team decision-making for challenging cancer cases.}
\newblock Proc. Mach. Learn. Res. {\bf 298}, 1-31 (2025).




\bibitem{B22}
Chen D, Parsa R, Swanson K, Nunez J, et al.
\newblock {Large language models in oncology: a review.}
\newblock BMJ Oncol. {\bf 4}, 000759 (2025).

\bibitem{B23}
Li C, Kalisa A, Roohani S, et al.
\newblock {The initation game: large language models versus multidisciplinary tumor boards: benchmarking AI against 21 sarcoma centers from the ring trail.}
\newblock  J. Cancer Res. Clin. Oncol. {\bf 151}, 248 (2025).

\bibitem{B24}
Benary M, Wang X, Schmidt M, Soll D, et al.
\newblock {Leverage large language models for decision support in personalized oncology.}
\newblock  JAMA Net. Open {\bf 6}, e2343689 (2023).


\bibitem{B26}
Sheng G, Zhang C, Ye Z, et al.
\newblock {HybridFlow: A Flexible and Efficient RLHF Framework.}
\newblock  arXiv: 2409.19256,  (2024).

\bibitem{B27}
Hu E, Shen Y, Wallis P, et al.
\newblock {LoRA: Low-rank adaptation of large language models.}
\newblock  arXiv:2106.09685v2,  (2021).

\bibitem{B28}
DeepSeek-AI.
\newblock {DeepSeek-R1: incentivizing reasoning capability in LLMs via reinforcement learning.}
\newblock  arXiv:2501.12948v1,  (2025).


\end{thebibliography}
\end{document}